\documentclass{article} 
\usepackage{nips12submit_e,times}
\usepackage{times}
\usepackage{epsfig}
\usepackage{graphicx}
\usepackage{amsmath}
\usepackage{amssymb}

\title{Submodularity of a Set Label Disagreement Function}

\author{
Toufiq Parag \\
Janelia Farm Research Campus- HHMI\\
Ashburn, VA 20147 \\
\texttt{paragt@janelia.hhmi.org} \\
}

\nipsfinalcopy 

\begin{document}

\maketitle

\begin{abstract}
A set label disagreement function is defined over the number of
variables that deviates from the dominant label. The dominant label
is the value assumed by the largest number of variables within a set
of binary variables. The submodularity of a certain family of set
label disagreement function is discussed in this manuscript. Such
disagreement function could be utilized as a cost function in
combinatorial optimization approaches for problems defined over
hypergraphs.
\end{abstract}

\section{Introduction}
Let $\mathbf{x}= \{ x_{1}, \dots, x_{k}\}$ be the binary labels
(i.e., $x_i \in \{0, 1\}$) of a set of datapoints $\mathbf{v}= \{
v_{1}, \dots, v_{k}\}$  of size $k$. The dominant label
$\rho({\mathbf{x}})$ among the labels of $\mathbf{x}$ is defined as
the label that largest number of vertices are assigned to. For
example, $\rho(\{1, 0, 1, 1\}) = 1$ and $\rho(\{0, 0, 0, 1\}) = 0$.
Let us also denote the number of variables to assume a value $c \in
\{0, 1\}$ by $n_c$.

In what follows, we analyze what label disagreement functon $g$,
defined on $k-n_{\rho(\mathbf{x})}$, are submodular. Such function
can be utilized in combinatorial approach of hypergraph
clustering~\cite{nagano10clustering} where the disagreement function
$g$ acts as a penalty for number of hyperedge nodes that deviates
from the dominant label $\rho(\mathbf{x})$. It may also be exploited
as a cost function for Markov Random Fields (MRF) with higher order
potentials~\cite{Freedman05energyminimization}.

Unlike the method presented in Kolmogorov and
Zabih~\cite{kolmogorov-04} for proving submodularity of functions
defined over subsets larger than 2, \emph{we do not project} the
sets on to pairs of variables and show them to be submodular. The
following analysis provides the researchers an alternate approach to
do the same exploiting the label arrangement of $\mathbf{v}$
directly.


\section{Proposition}\label{A:HYPERCUTSUBMOD} For
nondecreasing concave $g$, the label disagreement function
$d(\mathbf{x})= g(k - n_{\rho({\mathbf{x}})})$ is submodular.

\vspace{0.2cm} \noindent \textbf{Proof:}



Let $\mathbf{a} = [a_{1}, \dots,a_{k}]^T$ and $ \mathbf{b} = [b_{1},
\dots, b_{k}]^T $  be two instantiation of the labels $\mathbf{x}.$
Denoting $\vee$ and $\wedge$ as element-wise logical \lq or\rq and~
\lq and\rq~ respectively, we need to prove the following for the
submodularity of $d$.

\vspace{-0.2cm} {
\begin{equation}\label{E:PROOF2COND}
\small d(\mathbf{a}) + d(\mathbf{b}) \ge  d(\mathbf{a}
\vee \mathbf{b} ) + d(\mathbf{a}\wedge \mathbf{b})
\vspace{-0.4cm}
\end{equation}}


Table~\ref{T:PROOF2} describes the possible configuration of values
in $\mathbf{a}$ and $\mathbf{b}$. The first two rows of
Table~\ref{T:PROOF2} states that, there are $\kappa_1$ zeros among
the values of both $\mathbf{a}$ and $\mathbf{b}$;  and there are
$\kappa_4$ ones in both of them. But, values of  $\mathbf{a}$ and
$\mathbf{b}$ differs in $\kappa_2+\kappa_3$ places, i.e., there are
$\kappa_2+\kappa_3$ places where $a_i = 1- b_i$. The rows of $a\vee
b$ and $a\wedge b$ in Table~\ref{T:PROOF2} show the resulting
configuration due to the values in $\mathbf{a}$ and $\mathbf{b}$.

\begin{table}[h]
\small
\begin{center}
\begin{tabular}{c|c|c|c|c}
  \hline
  config & $\kappa_1$ & $\kappa_2$ & $\kappa_3$ & $\kappa_4$ \\
  \hline
  $\mathbf{a}$ & 0 & 0 & 1 & 1 \\
  $\mathbf{b}$ & 0 & 1 & 0 & 1 \\
  $\mathbf{a} \vee \mathbf{b}$ & 0 & 1 & 1 & 1 \\
  $\mathbf{a} \wedge \mathbf{b}$ & 0 & 0 & 0 & 1 \\
  \hline
\end{tabular}
\caption{\footnotesize Possible combinations of the values in
$\mathbf{a}$ and $\mathbf{b}.$ The first row of the table implies
that $\kappa_1 + \kappa_2$ values of $\mathbf{a}$ are zeros and
$\kappa_3+\kappa_4$ of them are ones.} \label{T:PROOF2}
\end{center}
\end{table}

Let us examine all possible cases of $\rho(\mathbf{a} \vee
\mathbf{b})$ and $\rho(\mathbf{a} \wedge \mathbf{b})$ values using
Table~\ref{T:PROOF2} and prove that the condition in
(\ref{E:PROOF2COND}) holds for them.
\begin{itemize}
  \item \textbf{Case $\rho(\mathbf{a} \vee
\mathbf{b}) = 0$ :} According to Table~\ref{T:PROOF2}, this case
enforces that $\rho(\mathbf{a} \wedge \mathbf{b}) = 0$.
Therefore, we need to show the following for
(\ref{E:PROOF2COND}) to hold.
\begin{align}\label{E:PROOF2CASE1}
& g(\kappa_3 + \kappa_4) + g(\kappa_2+\kappa_4) \ge
g(\kappa_2+\kappa_3+\kappa_4) + g(\kappa_4) \nonumber \\
\Rightarrow~~& {g(\kappa_2+\kappa_4)- g(\kappa_4) \over \kappa_2}
\ge {g(\kappa_2+\kappa_3+\kappa_4)- g(\kappa_3 + \kappa_4) \over
\kappa_2}.
\end{align}

\noindent This condition holds only as $g$ is a nondecreasing
concave function (i.e., nonincreasing slope).

  \item \textbf{Case $\rho(\mathbf{a} \wedge \mathbf{b}) = 1$
:} According to Table~\ref{T:PROOF2}, this case enforces that
$\rho(\mathbf{a} \vee \mathbf{b}) = 1$ and the proof is similar
to that of above case.

  \item \textbf{Case $\rho(\mathbf{a} \vee
\mathbf{b}) = 1$ and $\rho(\mathbf{a} \wedge \mathbf{b}) = 0$ :}
From Table~\ref{T:PROOF2}, we can write $d(\mathbf{a} \vee
\mathbf{b}) = g(\kappa_1)$ and $d(\mathbf{a} \wedge \mathbf{b})
= g(\kappa_4)$. It is straightforward to show that if
$\rho(\mathbf{a}) \ne \rho(\mathbf{b})$, the inequality
(\ref{E:PROOF2COND}) holds due to the nondecreasing nature of
$g.$

If we have $\rho(\mathbf{a}) = \rho(\mathbf{b}) = 0$, the
condition we need to satisfy is as follows.

\begin{equation}\label{E:PROOF2CASE3}
g(\kappa_3 + \kappa_4) + g(\kappa_2+\kappa_4) \ge g(\kappa_1) +
g(\kappa_4).
\end{equation}

The concavity of $g$ gives us,
\begin{align}
g(\kappa_2 + \kappa_4) + g(\kappa_3+\kappa_4) \ge
g(\kappa_2+\kappa_3+\kappa_4) + g(\kappa_2 + \kappa_4)
\end{align}

We know that $\rho(\mathbf{a} \vee \mathbf{b}) = 1$ implies
$\kappa_2+ \kappa_3+\kappa_4 \ge \kappa_1 $. Furthermore, due to
nondecreasing nature of $g$, we have
$g(\kappa_2+\kappa_3+\kappa_1) \ge g(\kappa_1)$. Therefore
inequality in (\ref{E:PROOF2CASE3}) holds and $d$ is submodular.
Similar proof can be reproduced for $\rho(\mathbf{a}) =
\rho(\mathbf{b}) = 1$.

  \item \textbf{Case $\rho(\mathbf{a} \vee
\mathbf{b}) = 0$ and $\rho(\mathbf{a} \wedge \mathbf{b}) = 1$ :}
For this case to occur we need $\kappa_1 \ge
\kappa_2+\kappa_3+\kappa_4$ and $\kappa_4 \ge \kappa_1 +
\kappa_2 + \kappa_3$. These conditions will only be true when
$\kappa_2+ \kappa_3 = 0$ which implies $\kappa_1 = \kappa_4.$
All the possible scenarios can be proved trivially with
$\kappa_1$ being equal to $\kappa_4$.  $\square$
\end{itemize}

{\small
\bibliographystyle{unsrt}
\bibliography{submodular_label}
}

%
%
%
%
%

\end{document}